# Deep Convolutional Neural Network for Age Estimation based on VGG-Face Model

Zakariya Qawaqneh(1), Arafat Abu Mallouh(1), Buket D. Barkana(2)
(1)Department of Computer Science and Engineering, University of Bridgeport,
(2)Department of Electrical Engineering, University of Bridgeport,
Technology Building, Bridgeport CT 06604 USA
Emails: {zqawaqneh; aabumall@my.bridgeport.edu}, bbarkana@bridgeport.edu

*Abstract*— Automatic age estimation from real-world and unconstrained face images is rapidly gaining importance. In our proposed work, a deep CNN model that was trained on a database for face recognition task is used to estimate the age information on the Adience database. This paper has three significant contributions in this field. (1) This work proves that a CNN model, which was trained for face recognition task, can be utilized for age estimation to improve performance; (2) Over fitting problem can be overcome by employing a pretrained CNN on a large database for face recognition task; (3) Not only the number of training images and the number subjects in a training database effect the performance of the age estimation model, but also the pre-training task of the employed CNN determines the model's performance.

*Index Terms*—Age classification, estimation, Convolutional Neural Network, face images, unconstrained database.

## I. INTRODUCTION

Recently, many applications from biometrics, security control to entertainment use the information extracted from face images that contain information about age, gender, ethnic background, and emotional state. Automatic age estimation from facial images is one of the popular and challenging tasks that have different fields of applications such as controlling the content of the watched media depending on the customer's age [1-3].

Automatic estimation of the age is a challenging process since the aging process among humans is non-uniform. In addition, extracting an effective feature set from a 2D image for age estimation is another challenge to overcome.

CNNs showed significant success in face recognition, image classification, and object recognition. It consists of different convolutional layers where each layer processes the output of the previous layer in order to produce a robust and compact output. CNNs can be described as deep networks if the number of layers inside the network is relatively a large number. If a CNN is characterized as a deep network, hence a large database is needed to optimize the parameters during the training process [4].

In this work, a pretrained deep CNN is utilized to estimate the age from unconstrained face images. The deep network was trained initially for face recognition task on a large database other than the one used for age estimation. Deep CNNs are capable to perform tasks efficiently with a condition in which they are trained on large databases. Currently, there is no relatively large database for age estimation task.

## II. LITERATURE REVIEW

Earliest methods used the size and proportions on the human face. These methods are limited to young ages due to the nature of the human head that does change significantly in the adulthood [5]. Later, the active appearance model (AAM) [6], ages pattern subspace (AGES) [7, 8], age manifold [9, 11], 3D morphable model [10] are proposed. Semantic-level description model for facial features characterization was used in [12, 13]. The effective texture descriptor and local binary patterns (LBP) were used by [14] for appearance feature extraction.

Table I summarizes the other related works. The previously mentioned literature utilized relatively small databases. Moreover, the type of the face images was lab and constrained images. Recently, new benchmarks for age estimation using unconstrained facial images have been designed. The unconstrained face images increase the challenge of age estimation due to the high variation of the real-world images which are taken in different environments.

TABLE I
PREVIOUS AGE ESTIMATION METHODS

| Work | Feature set | Classifier | Accuracy | Database |
|---|---|---|---|---|
| [15] | Gabor feature | Fuzzy-LDA | 91% | Private |
| [16, 17] | Spatial Flexible Patch | GMM | 4.94 MAE | FG-NET |
| [18] | Graphical facial features, topology, geometry | ANN | 5.974 MAE | FG-NET |
| [19] | Bio-inspired features (BIF) | SVM | 4.77 MAE | FG-NET |
| [20] | BIF and the age manifold | SVM | 2.61 MAE - f  2.58 MAE - m | YGA |

Two examples of recent benchmarks are the Group Photos [21] and the Adience benchmark [22]. The Adience is considered as the newest and the most challenging benchmark for age estimation using face images. The studies [2, 23, 24] were evaluated on the Alience.

## III. CNNs ARCHITECTURE AND TRAINING FOR AGE ESTIMATION

In this section we explain our proposed deep CNN that estimates the age of a subject from 2D images. Deep CNNs are powerful models, which are capable to capture effective information. One of the most challenging problems in the machine learning is the overfitting problem that occurs when using small databases. In deep neural networks, the problem of the overfitting becomes even worse due to the fact that deep networks have millions of parameters, since they have several number of layers with thousands of nodes. All databases that are built for age classification and prediction are relatively small in size. They are not comparable in size with other databases designed for face recognition and image classification tasks. To overcome the overfitting problem, we architect our proposed deep CNN for age estimation by using a deep CNN model trained for face recognition on a very large database.

*A: Architecture*

Using a trained CNN as a facial feature extractor is expected to be useful as a keystone for training CNN to estimate the age from the face images. The proposed CNN architecture relies on a very deep face recognition CNN architecture which is capable of extracting facial features distinctively and robustly. As well as, it will be less prone to overfitting.

There are a few CNN models that were successfully trained for face recognition task. In this work, we use the VGG-face model proposed by [25] which achieved the state-of-the-art results on the LFW [26] and YFT [27] databases. VGG-Face consists of 11 layers, eight convolutional layers and 3 fully connected layers. As shown in Table II, each convolutional layer is followed by a rectification layer, whereas a max pool layer is operated at the end of each convolutional block.

TABLE II
ARCHITECTURE AND CONFIGURATION OF THE PROPOSED MODEL

| Layer# | 0 | 1 | 2 | 3 | 4 |
|---|---|---|---|---|---|
| Name | n/a | Conv1 | Relu1 | Norm1 | Pool1 |
| Support | n/a | 3 | 1 | 3 | 1 |
| Filt dim | n/a | 3 | n/a | 64 | n/a |
| Num filts | n/a | 64 | n/a | 64 | n/a |
| Stride | n/a | 1 | 1 | 1 | 1 |
| pad | n/a | 1 | 0 | 1 | 0 |

| Layer# | 5 | 6 | 7 | 8 | 9 |
|---|---|---|---|---|---|
| Name | Conv2 | Relu2 | Norm2 | Pool2 | Conv3 |
| Support | 2 | 3 | 1 | 3 | 1 |
| Filt dim | n/a | 64 | n/a | 128 | n/a |
| Num filts | n/a | 128 | n/a | 128 | n/a |
| Stride | 2 | 1 | 1 | 1 | 1 |
| pad | 0 | 1 | 0 | 1 | 0 |
| Layer# | 10 | 11 | 12 | 13 | 14 |
| Name | Relu3 | Conv4 | Relu4 | Conv5 | Relu3_2 |
| Support | 2 | 3 | 1 | 3 | 1 |
| Filt dim | n/a | 128 | n/a | 256 | n/a |
| Num filts | n/a | 256 | n/a | 256 | n/a |
| Stride | 2 | 1 | 1 | 1 | 1 |
| pad | 0 | 1 | 0 | 1 | 0 |
| Layer# | 15 | 16 | 17 | 18 | 19 |
| Name | Conv3_3 | Relu3_3 | Mpool3 | Conv4_1 | Relu4_1 |
| Support | 3 | 1 | 2 | 3 | 1 |
| Filt dim | 256 | n/a | n/a | 256 | n/a |
| Num filts | 256 | n/a | n/a | 512 | n/a |
| Stride | 1 | 1 | 2 | 1 | 1 |
| pad | 1 | 0 | 0 | 1 | 0 |
| Layer# | 20 | 21 | 22 | 23 | 24 |
| Name | Onv4_2 | Relu4_2 | Conv4_3 | Relu4_3 | Mpool4 |
| Support | 3 | 1 | 3 | 1 | 2 |
| Filt dim | 512 | n/a | 512 | n/a | n/a |
| Num filts | 512 | n/a | 512 | n/a | n/a |
| Stride | 1 | 1 | 1 | 2 | 2 |
| pad | 1 | 0 | 1 | 0 | 0 |
| Layer# | 25 | 26 | 27 | 28 | 29 |
| Name | Conv5_1 | Relu5_1 | Conv5_2 | Relu5_2 | Conv5_3 |
| Support | 3 | 1 | 3 | 1 | 3 |
| Filt dim | 512 | n/a | 512 | n/a | 512 |
| Num filts | 512 | n/a | 512 | n/a | 512 |
| Stride | 1 | 1 | 1 | 1 | 1 |
| pad | 1 | 0 | 1 | 0 | 1 |
| Layer# | 30 | 31 | 32 | 33 | 34 |
| Name | relu5_3 | pool5 | fc6 | relu6 | drop6 |
| Support | 1 | 2 | 7 | 1 | 1 |
| Filt dim | n/a | n/a | 512 | n/a | n/a |
| Num filts | n/a | n/a | 4096 | n/a | n/a |
| Stride | 1 | 2 | 1 | 1 | 1 |
| pad | 0 | 0 | 0 | 0 | 0 |
| Layer# | 35 | 36 | 37 | 38 | 39 |
| Name | fc7 | relu7 | drop7 | fc8 | relu8 |
| Support | 1 | 1 | 1 | 1 | 1 |
| Filt dim | 4096 | n/a | n/a | 5000 | n/a |
| Num filts | 5000 | n/a | n/a | 5000 | n/a |
| Stride | 1 | 1 | 1 | 1 | 1 |
| pad | 0 | 0 | 0 | 0 | 0 |
| Layer# | 40 | 41 | 42 | | |
| Name | drop8 | fc9 | prob | | |
| Support | 1 | 1 | 1 | | |
| Filt dim | n/a | 5000 | n/a | | |
| Num filts | n/a | 8 | n/a | | |
| Stride | 1 | 1 | 1 | | |
| pad | 0 | 0 | 0 | | |

The fully connected layers are a special case of the convolutional layers where the size of the filters and the input data are the same. The number of the input features for the first two fully connected layers is 4096. They are followed by a dropout layer with a drop rate of p=0.5.

The last fully connected layer of this model represents an N-way class predictor, where N represents the number of the labels (classes) in the database. In this work, we architect and retrain the VGG-Face model for age estimation by keeping the convolutional layers of this model unchanged while replacing the fully connected layers with four new fully connected layers. The first three fully connected layers are followed by dropout and rectification layers. The size of the first fully connected layer is 4096, while the size of the second and the third fully connected layers is 5000. The output size of the output layer represents the number of age labels which is 8.

*B: Training*

The input images are rescaled to 256 x 256 pixel, then randomly cropped to 224 x 224 pixel patches. The optimization of our proposed network is carried out by using stochastic gradient descent method with mini-batches of size 256 and momentum value of 0.9. In addition, the weight decay is set to 10-3. A dropout rate of 0.6 is used to regularize the network parameters during the training process. The training starts by 0.1 learning rate and then it is decreased by a factor of 10 whenever there is no improvement in the validation set accuracy result. The weights between the newly added fully connected layers are initialized by using a Gaussian distribution with zero mean and 10-2 standard deviation, while the biases are initialized to zero. The RGB input image is fed to the input layer of the network. Then the output of each hidden layer is fed to the next hidden layer as an input until the probability of the output layer (last layer) of the network is calculated.

The stochastic gradient decent method optimizes and finds the parameters of the connected layers that minimize the prediction of the softmax-log-loss for age estimation as shown in Fig. 1. In the meanwhile, the convolutional layers' parameter are not changed and kept frozen. In other words, we optimize the fully connected layers' parameters to predict the age of subjects while not changing the parameters of the convolutional layers which were trained and optimized for face recognition task by [25].

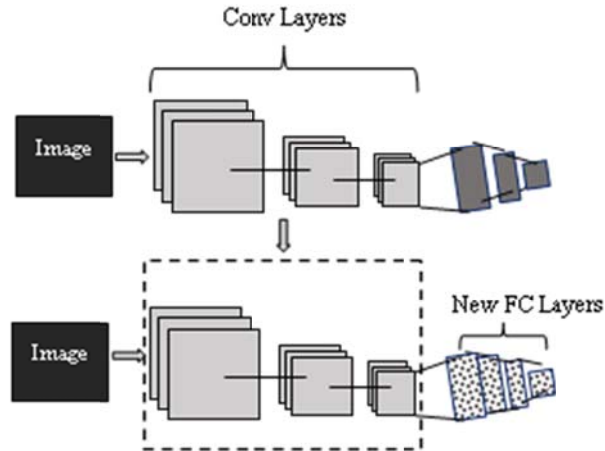

Fig. 1. The proposed model

*C: Prediction*

A test image is rescaled to 256x256 pixels. Then three images are extracted of size 224x224. The first image is obtained from the center of the original test image. The second and third images are extracted from the bottom-left and upper-right corners of the original test image, respectively. By using the trained CCN network, these three extracted images are fed to the model to calculate the softmax probability output vector for each image. To obtain the final probability vector of class scores for the original test image, the output score vectors of the three images are averaged. This method reduces the impact of poor quality images, such as low-resolutions, and occlusions.

## IV. EXPERIMENTAL RESULTS AND DISCUSSION

Different experiments have been carried out to evaluate the proposed work. The specifications of the used benchmark database and details about the conducted experiments are reported in this section.

The Adience benchmark is one of the newest databases designed for age estimation from face images. In this work, the Adience is used to evaluate the efficiency of the proposed work. It consists of unconstrained face images of 2284 subjects and has 8 age labels which were uploaded to the Flicker website. Table III lists the age labels and the number of images per label for males and females.

TABLE III
THE ADIENCE DATABASE

| Gender | Labels in years | | | | | | | | Total |
|---|---|---|---|---|---|---|---|---|---|
| | 0-2 | 4-6 | 8-13 | 15-20 | 25-32 | 38-43 | 48-53 | 60- | |
| Female | 682 | 1234 | 1360 | 919 | 2589 | 1056 | 433 | 427 | 9411 |
| Male | 745 | 928 | 934 | 734 | 2308 | 1294 | 392 | 442 | 8192 |

A performance comparison between the proposed work and the state-of-the-art methods reported in [22] and [2] is presented in Table IV. Furthermore, Table IV shows the 1-off accuracy that represents the accuracy when the result is off by 1 adjacent age label left or right. Based on our results, the proposed work significantly outperforms the state-of-the-art results in terms of the exact accuracy and for 1-off accuracy. These results confirm the efficiency of the proposed work. Table V gives the confusion matrix for the proposed model.

TABLE IV
OVERALL ACCURACY FOR DIFFERENT CNN ARCHITECTURE

| Method | Exact Accuracy | 1-off Accuracy |
|---|---|---|
| [22] | 45.1 | 79.5 |
| [2] using single crop | 49.5 | 84.6 |
| [2] using over-sample | 50.7 | 84.7 |
| **Proposed Work** | **59.9** | **90.57** |

TABLE V
CONFUSION MATRIX FOR THE PROPOSED MODEL BY USING THE VGG-FACE CNN

| Actual / Predict | 0-2 | 4-6 | 8-13 | 15-20 | 25-32 | 38-43 | 48-53 | 60- |
|---|---|---|---|---|---|---|---|---|
| 0-2 | **93.17** | 6.42 | 0.21 | 0.00 | 0.21 | 0.00 | 0.00 | 0.00 |
| 4-6 | 26.84 | **62.11** | 7.37 | 1.93 | 1.58 | 0.00 | 0.18 | 0.00 |
| 8-13 | 1.76 | 6.18 | **42.06** | 12.94 | 35.59 | 0.29 | 1.18 | 0.00 |
| 15-20 | 1.76 | 0.44 | 4.41 | **24.23** | 64.76 | 0.00 | 4.41 | 0.00 |
| 25-32 | 0.00 | 0.09 | 0.85 | 3.22 | **86.17** | 3.31 | 4.83 | 1.52 |
| 38-43 | 0.39 | 0.20 | 0.39 | 1.78 | 59.76 | **8.88** | 22.88 | 5.72 |
| 48-53 | 0.00 | 0.00 | 0.00 | 0.00 | 19.50 | 8.71 | **38.17** | 33.61 |
| 60- | 0.00 | 0.00 | 0.00 | 1.17 | 1.95 | 1.17 | 35.02 | **60.70** |

From Table V, it is noticed that the 0-2 year old age label is estimated with the highest accuracy, 93.17%. Images of infants contain distinctive features that enable the classifier to distinguish this age group easily. 15-20 and 38-43 year old age labels classified with the lowest accuracies, 24.3% and 8.88%, respectively. Both labels are adjacent to Label 5 that is the 25-32 year old age group. Label 5 is classified with accuracy of 86.17%. These results might be a result of the difference in subject numbers between labels 4 and 6 and the label 5. The number of subjects in label 5 is very large compared to the labels 4 and 6. Therefore, label 5 is trained and classified better by affecting the results of its adjacent labels. Labels 4 and 6 are highly misclassified with label 5, 64.76%, 59.76%, respectively.

TABLE VI
OVERALL ACCURACY OF DIFFERENT CNN ARCHITECTURES (%)

| Label | Proposed model by VGG-Face CNN | Model by GoogLeNet CNN |
|---|---|---|
| 0-2 | 93.17 | 86.75 |
| 4-6 | 62.11 | 27.89 |
| 8-13 | 42.06 | 21.47 |
| 15-20 | 24.23 | 14.10 |
| 25-32 | 86.17 | 76.61 |
| 38-43 | 8.88 | 12.03 |
| 48-53 | 38.17 | 7.05 |
| 60- | 60.70 | 34.63 |
| **Overall Accuracy** | **59.90** | **45.07** |

To prove the efficiency of the proposed work furthermore, the GoogLeNet [28] model that was trained for image classification on ImageNet ILSVRC database [29] is retrained, fine-tuned and tested for age estimation by using the proposed model. We modified and fine-tuned the GoogLeNet CNN to perform age prediction by replacing the fully connected layers and changing the number of nodes. In the modified architecture, four fully connected layers are used where the number of nodes for each layer is 1024, 2048, 2048, and 8 respectively. Then the modified GoogLeNet is retrained and fine-tuned while preserving the convolutional layers unchanged during the training process as mentioned in section 3. The modified GoogLeNet CNN achieved 45.07% in age estimation. Table VI presents the performance of the proposed model by using VGG-Face CNN and the GoogLeNet CNN for age estimation.

To perform age estimation, the employment of the GoogLeNet CNN that was trained for image classification task provided reasonable results. However, based on the results in Table VI, it is clear that features extracted by using a CNN model trained for face recognition task are more effective than the features extracted by using a CNN model trained for image classification.

Fig.2 shows some of the images of the Adience database. Although these are very challenging images having blur, poor lighting, low-resolution, motion, pose, and facial expressions, they all are classified correctly using the proposed work.

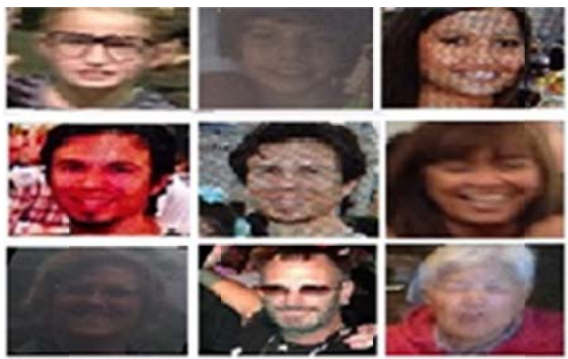

Fig. 2. Some of the successfully classified challenging images from the Adience database

## V. CONCLUSION

In this paper, we proposed a model to perform age estimation based on facial images by using a deep CNN called VGG-Face that was trained for face recognition on a large database. The VGG-Face CNN is modified and fine-tuned to perform age estimation. The proposed model outperforms the previous works by 9% on the Adience database which is the newest challenging age estimation benchmark that consists of unconstrained face images. The GoogLeNet was trained on a very large database that contain millions of training images, its performance in age estimation is not competitive with the proposed model using the VGG-Face. Not only the number of the training images and the number subjects in a training database effect the performance for age estimation, but also the pre-training task of the employed CNN determines the network's performance for age estimation.